\documentclass{article}

\usepackage[preprint]{neurips_2024}
\usepackage{hyperref}
\usepackage{amsfonts,amsmath,url,graphicx,booktabs}
\usepackage{subcaption}

\usepackage[utf8]{inputenc} 
\usepackage[T1]{fontenc}    
\usepackage{hyperref}       
\usepackage{url}            
\usepackage{booktabs}       
\usepackage{amsfonts}       
\usepackage{nicefrac}       
\usepackage{microtype}      
\usepackage{xcolor}         

\title{Red Skills or Blue Skills? A Dive Into Skills Published on ClawHub}

\author{%
  Haichuan Hu \\
  The Hong Kong Polytechnic University\\
  \texttt{haichuan.hu@connect.polyu.hk} \\
  \And
  Ye Shang \\
  Nanjing University\\
  \texttt{yeshang@smail.nju.edu.cn} \\
  \And
  Quanjun Zhang \\
  Nanjing University of Science and Technology \\
  \texttt{quanjunzhang@njust.edu.cn} \\
}

\begin{document}

\maketitle

\begin{abstract}
Skill ecosystems have emerged as an increasingly important layer in Large Language Model (LLM) agent systems, enabling reusable task packaging, public distribution, and community-driven capability sharing. However, despite their rapid growth, the functionality, ecosystem structure, and security risks of public skill registries remain underexplored. In this paper, we present an empirical study of ClawHub, a large public registry of agent skills. We build and normalize a dataset of 26,502 skills, and conduct a systematic analysis of their language distribution, functional organization, popularity, and security signals. Our clustering results show clear cross-lingual differences: English skills are more infrastructure-oriented and centered on technical capabilities such as APIs, automation, and memory, whereas Chinese skills are more application-oriented, with clearer scenario-driven clusters such as media generation, social content production, and finance-related services. We further find that more than 30\% of all crawled skills are labeled as suspicious or malicious by available platform signals, while a substantial fraction of skills still lack complete safety observability. To study early risk assessment, we formulate submission-time skill risk prediction using only information available at publication time, and construct a balanced benchmark of 11,010 skills. Across 12 classifiers, the best Logistic Regression achieves a accuracy of 72.62\% and an AUROC of 78.95\%, with primary documentation emerging as the most informative submission-time signal. Our findings position public skill registries as both a key enabler of agent capability reuse and a new surface for ecosystem-scale security risk.

\parbox{0.033\textwidth}{\includegraphics[width=\linewidth]{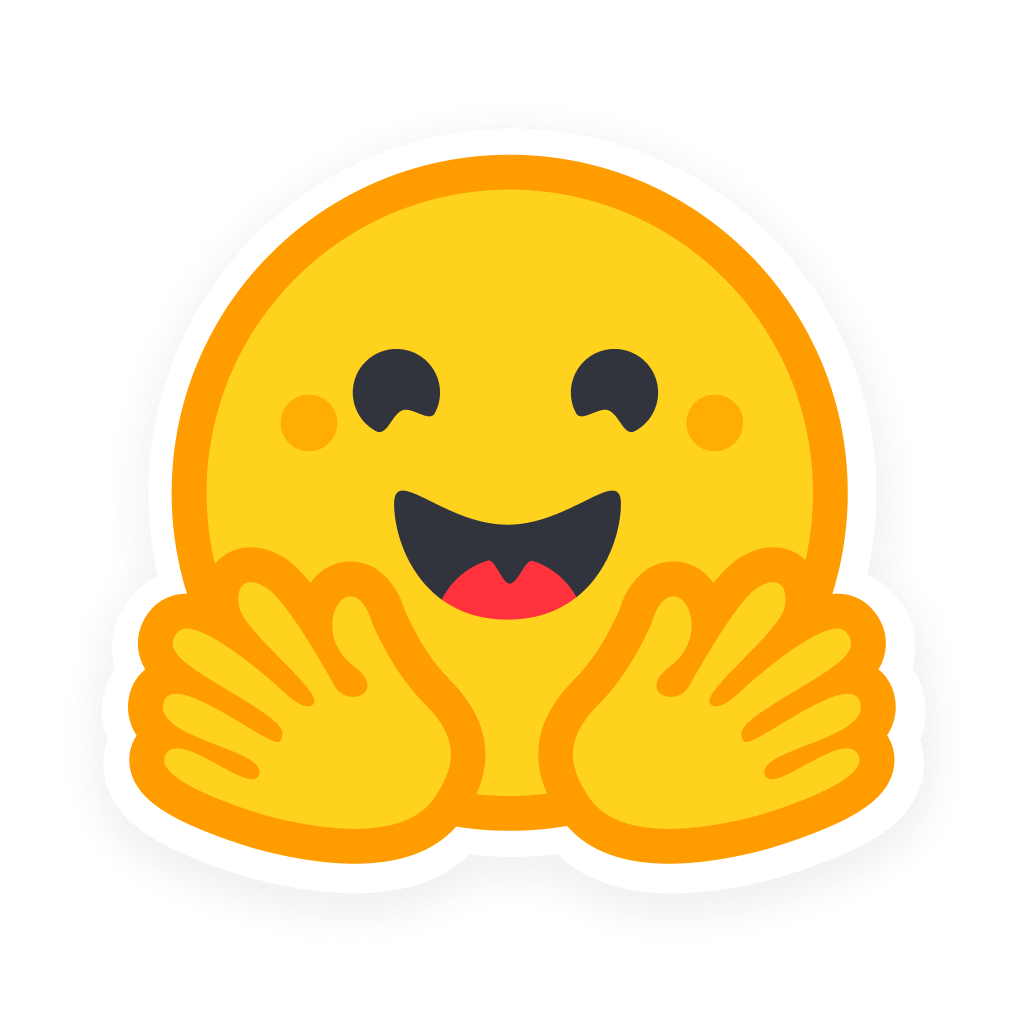}}\hspace{0.5mm}%
\href{https://huggingface.co/datasets/tomhu/ClawSkills}{{\texttt{https://huggingface.co/datasets/tomhu/ClawSkills}}}

\parbox{0.033\textwidth}{\includegraphics[width=\linewidth]{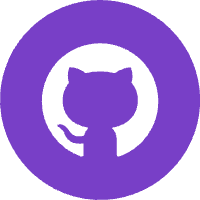}}\hspace{0.5mm}%
\href{https://github.com/Tomsawyerhu/Clawhub-Skills-Analysis}{{\texttt{https://github.com/Tomsawyerhu/Clawhub-Skills-Analysis}}}

\end{abstract}

\section{Introduction}

As Large Language Models (LLMs) evolve from static question-answering systems~\citep{bahak2023evaluating,wu2023towards,chan2023case,shen2023chatgpt} into agents~\citep{xi2025rise,yang2025comprehensive,guo2024large} capable of invoking external resources and executing complex tasks, mechanisms for encapsulating and reusing model capabilities have also developed rapidly. In this process, skills~\citep{ling2026agent,li2026skillsbench,chen2026skills} have gradually emerged as an important intermediate layer connecting foundation models with real-world tasks. Unlike atomic tools~\citep{qin2024tool,gou2023tora,singh2025agentic}, a skill is typically oriented toward a more complete task objective and encapsulates documentation, configuration, invocation logic, file assets, and one or more external capabilities, thereby offering stronger reusability, composability, and ease of deployment.

The development of skills has undergone a clear three-stage evolution. In the first stage, capability reuse was mainly realized through prompt templates~\citep{wang2025prompt,giray2023prompt} and plugin mechanisms~\citep{panda2024revolutionizing}, where reusable behaviors were introduced in relatively ad hoc and loosely structured ways. In the second stage, the advent of function calling~\citep{wang2025function,liu2024toolace}, tool use~\citep{lu2025toolsandbox,shen2024llm}, agent frameworks~\citep{xi2025rise}, and MCP~\citep{wang2025mcp,ahmadi2025mcp,sarkar2025survey} transformed these capabilities into more structured, callable, and composable execution units. In the third stage, community-oriented skill hubs and registries~\citep{liang2026skillnet,he2026openclaw} further turned skills into publishable, installable, shareable, and distributable software ecosystem units. Together, these stages have significantly lowered the barrier to building complex capabilities, accelerated the proliferation of agent applications, and stimulated the growth of skill-centered community ecosystems, content production, and task automation.

\begin{figure}[htbp]
\centering
\includegraphics[width=1.0\linewidth]{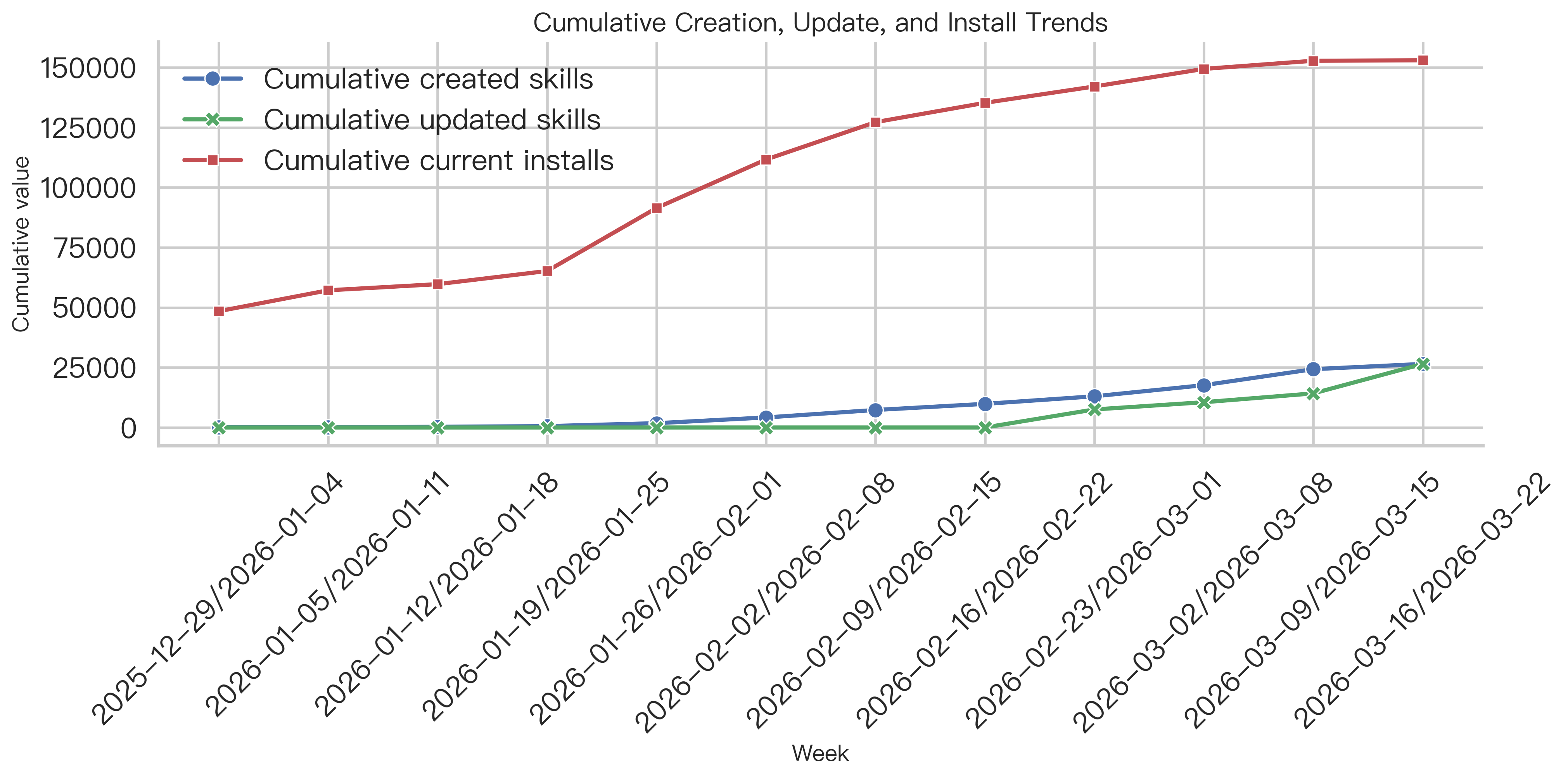}
\caption{Weekly accumulated creation, update and install on ClawHub.}
\label{fig:current_trend}
\end{figure}

This trajectory is becoming more and more visible. In October 2025, Anthropic officially published Claude Skills as reusable folders of instructions, scripts, and resources that improve task-specific performance, further formalizing skills as a practical abstraction for specialized agent behavior~\citep{anthropic2025lifesciences}. More recently, open community ecosystems (e.g., OpenClaw~\citep{openclaw2026skills}) have pushed this abstraction beyond local customization into large-scale public distribution. 
For example, through ClawHub~\citep{openclaw2026clawhub}, which is introduced by OpenClaw, skills are no longer merely private workflow components, but versioned, searchable, installable, and publicly shared artifacts that circulate within an open registry.  As shown in Figure~\ref{fig:current_trend}, ClawHub attracted substantial attention after its launch. Within just three months, the cumulative number of installations exceeded 150,000, and the growth rate continued to accelerate. This shift marks an important transition: skills are becoming not only execution units for agents, but also social and ecosystem-level objects that can accumulate visibility, reputation, downloads, and downstream influence.

However, the same properties that make skills easy to discover, reuse, and disseminate also introduce a new security surface. Unlike static prompts or isolated tools, skills often combine natural-language instructions, executable logic, external dependencies, and platform-level distribution channels, making their risks harder to characterize and contain. More importantly, once skills are embedded in public hubs such as ClawHub~\citep{openclaw2026clawhub}, their social attributes, including search exposure, ranking signals, stars, downloads, author identity, and community sharing, can amplify both adoption and harm. In such settings, unsafe or malicious skills are not merely individual artifacts; they can spread through visibility and trust mechanisms built into the ecosystem itself. Recent reports~\citep{tomshardware2026clawhub,theverge2026openclaw} on ClawHub have already shown that open skill registries can become channels for abusive or malicious submissions, highlighting the need to systematically study skill security, risk patterns, and early-stage detection in socially distributed registries.

In this paper, we have two main objectives. First, we aim to provide a systematic understanding of the current state of public skill ecosystems and to offer empirical insights that can inform their future development, governance, and standardization. Second, we focus on the security risks of public skill registries, seeking to understand how risky or malicious skills emerge, manifest, and spread in open ecosystems. To this end, we conduct an empirical study of 26,502 publicly released skills on ClawHub collected up to March 18, 2026, examining them from multiple perspectives, including skill content, metadata, and platform-level exposure and dissemination signals, and further exploring automated methods for early-stage risk detection on 11,010 filtered high-quality samples. Rather than treating unsafe skills as isolated artifacts, this work takes an ecosystem view and investigates how public distribution, search visibility, and trust mechanisms can amplify localized security issues into broader platform-level risks. To support future research, we also curate and release the collected dataset as a benchmark for skill ecosystem measurement, security analysis, and risk detection. Through this study, we aim to provide both empirical foundations and practical guidance for building safer, more trustworthy, and more sustainable open skill ecosystems.

\section{Benchmark}
\subsection{Data Collection}

\begin{figure}[htbp]
\centering
\includegraphics[width=1.0\linewidth]{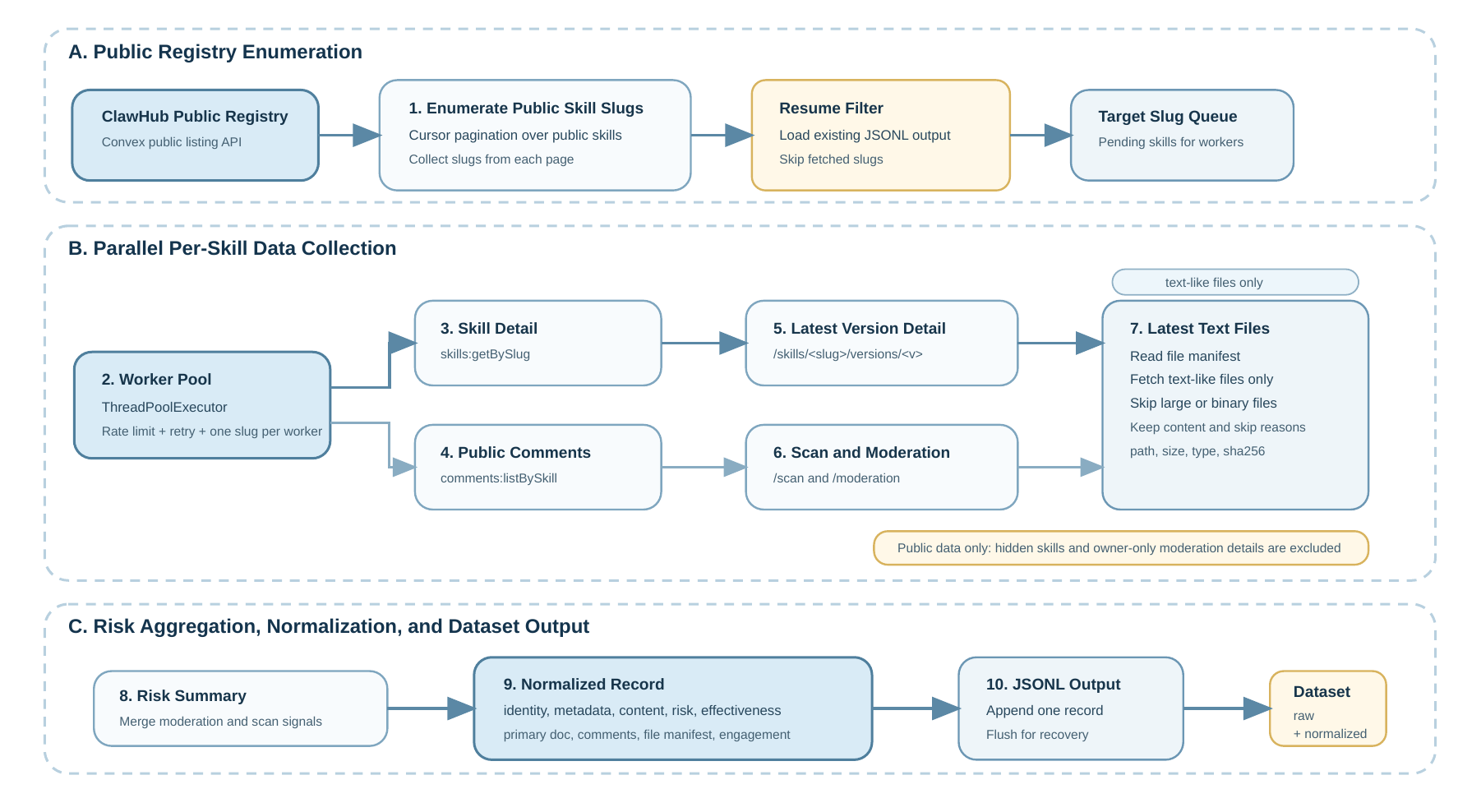}
\caption{Data collection pipeline.}
\label{fig:data_collection}
\end{figure}

To construct our dataset, we implemented a public-data crawler for ClawHub that systematically collects skill-level information from the platform and normalizes it into a unified data format. As shown in Figure~\ref{fig:data_collection}, the crawler operates in three stages. 
In the first stage, it enumerates publicly available skills through the registry listing interface using cursor-based pagination and records each skill slug as a crawl target. In the second stage, it performs parallel requests for each target skill to retrieve multiple categories of public information, including registry metadata, latest version metadata, public comments, scan results, moderation signals when available, and the contents of text-like files in the latest released version. During this process, non-text files and oversized files are skipped to reduce noise and bandwidth overhead, while file metadata and skip reasons are retained. In the third stage, the crawler aggregates security-related signals from scan and moderation endpoints, extracts structured content features such as primary documentation, file manifests, directory structure, and comment bodies, and converts each raw record into a normalized representation containing identity, metadata, content, risk, and effectiveness fields. Finally, each skill is written incrementally as one output record.

\subsection{Dataset Format}
Each skill record in our dataset contains five major components: \textit{identity}, \textit{metadata}, \textit{content}, \textit{risk}, and \textit{effectiveness}. The \textit{identity} field includes the skill slug, requested and resolved slugs, skill ID, owner information, and the latest version number. The \textit{metadata} field captures descriptive and platform-level attributes such as display name, summary, tags, creation and update timestamps, license information, and usage statistics. The \textit{content} field summarizes the observable skill contents, including primary documentation, file manifests, directory structure, text-like file contents, code-related signals, example files, and public comment bodies. The \textit{risk} field aggregates security-relevant signals collected from scan and moderation endpoints, including risk labels, reason codes, warning indicators, suspicious or malware-blocked flags, scan status, moderation status, and scanner outputs. Finally, the \textit{effectiveness} field provides a derived summary of publicly visible quality and utility signals, such as documentation completeness, content coverage, engagement statistics, version information, and an overall heuristic effectiveness score. Together, these fields provide a unified representation of each skill from functional, structural, social, and security perspectives, enabling downstream analysis, benchmarking, and risk detection.

\section{Experiments}

Based on the dataset we constructed, we design two downstream tasks: function analysis and risk identification. Function analysis mainly focuses on the overall development trends of skills, their functional distribution, and user preference patterns, while risk identification focuses on the potential vulnerabilities, hidden risks, and security issues that may exist in skills.

\subsection{Function Analysis}
When analyzing the functional characteristics of existing skills, we used the full dataset (26502 samples) as the analysis object and sequentially examined the skills' domain distribution, functional distribution, and download distribution.

\begin{figure}[htbp]
\centering
\includegraphics[width=1.0\linewidth]{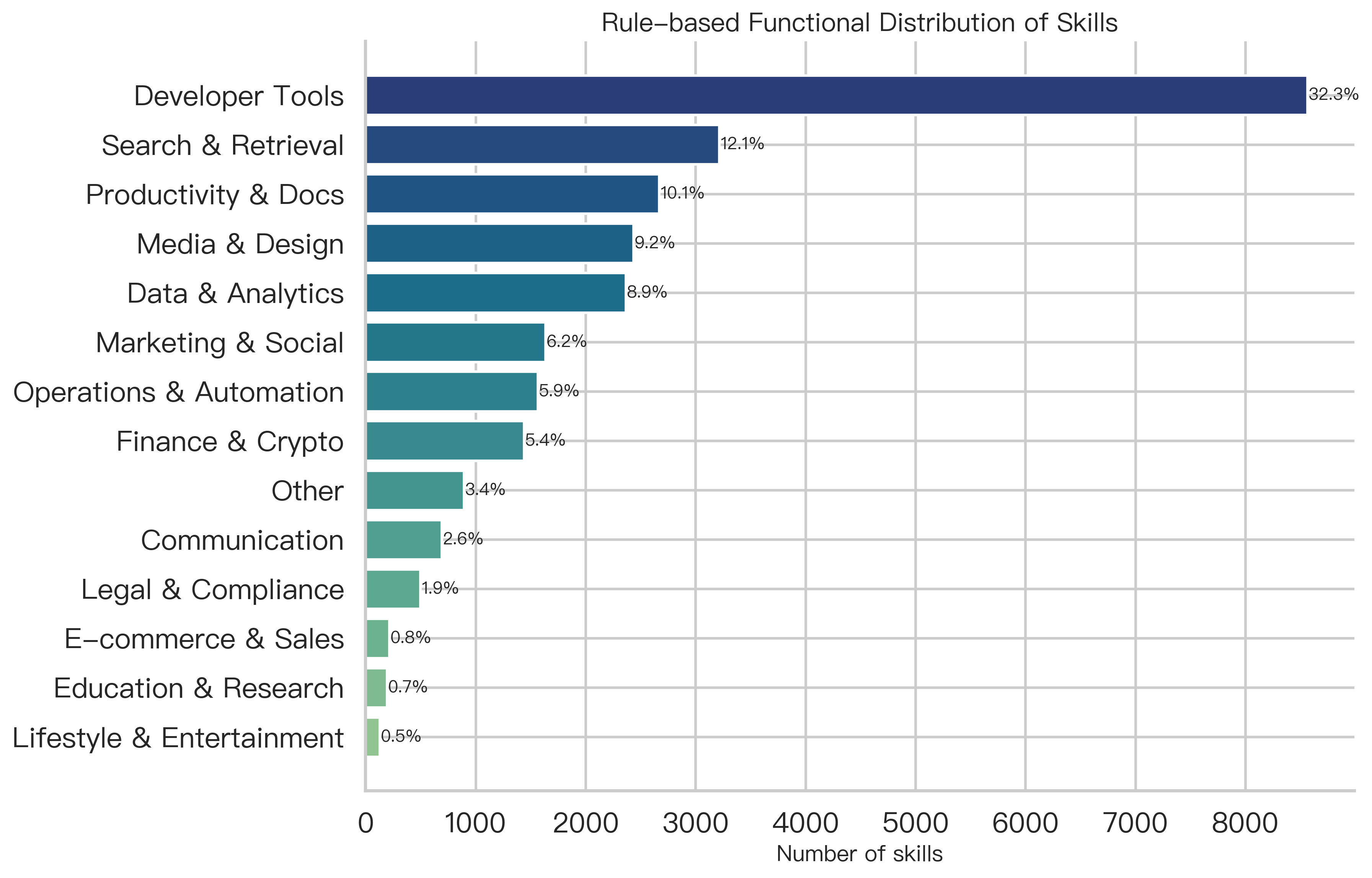}
\caption{Domain distribution of skills.}
\label{fig:function_distribution}
\end{figure}

As shown in Figure~\ref{fig:function_distribution}, we collected representative keywords for a set of common application domains and used a rule-based matching approach to identify the domain affiliation of each skill. The results show that more than 30\% of all skills belong to the \textit{Developer Tools} category, which is substantially higher than the second-largest category at 12.1\%. In addition, skills related to \textit{Product}, \textit{Design}, and \textit{Marketing} also account for considerable shares of the ecosystem, representing 10.1\%, 9.2\%, and 6.2\% of all skills, respectively. These findings suggest that the current skill ecosystem is strongly shaped by practical, task-oriented demands, with development-related scenarios occupying a dominant position. At the same time, the substantial presence of product-, design-, and marketing-oriented skills indicates that the ecosystem is not limited to technical users, but is gradually expanding toward broader professional and commercial application settings.

\begin{figure}[htbp]
    \centering
    \begin{subfigure}[t]{0.49\linewidth}
        \centering
        \includegraphics[width=\linewidth]{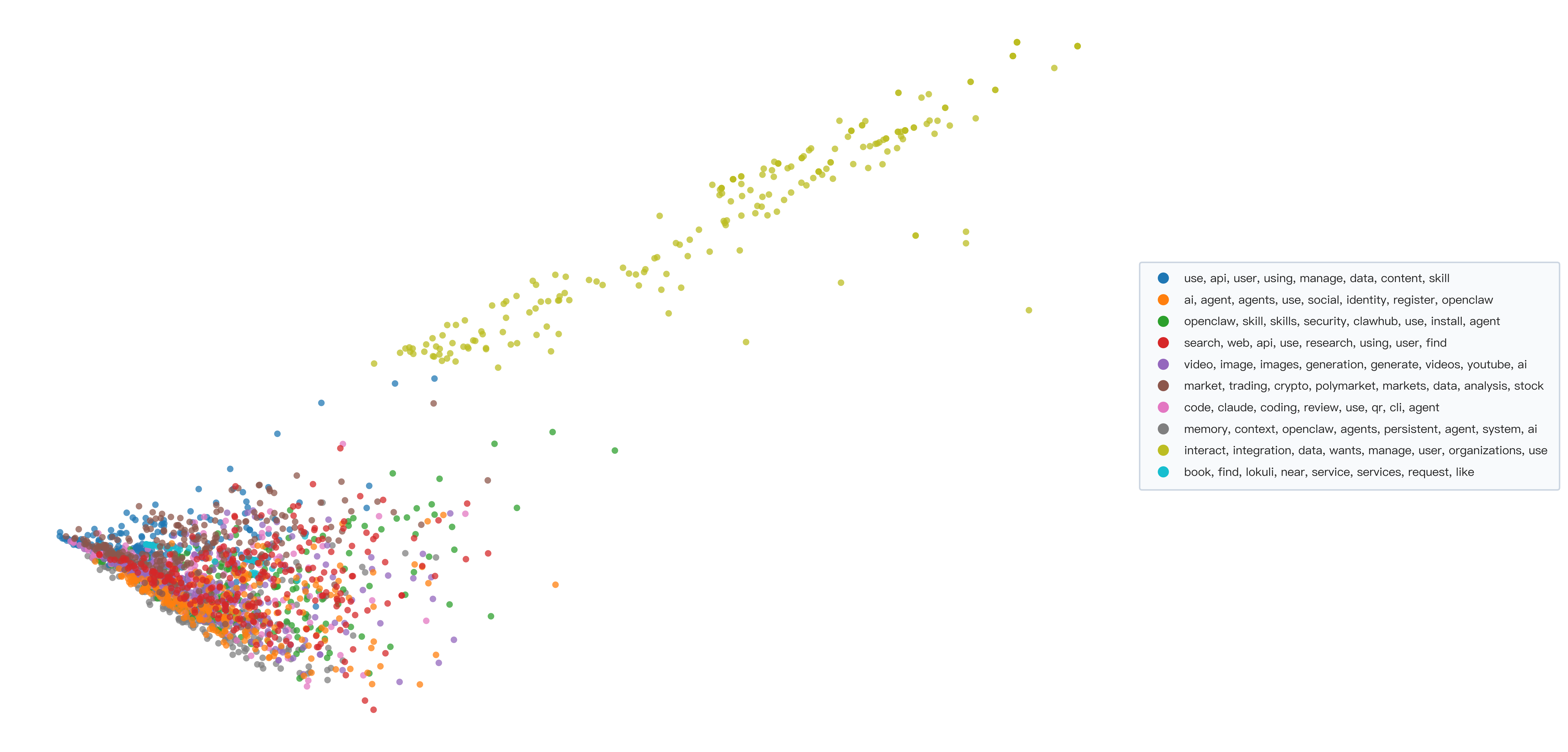}
        \caption{English skills (number of clusters = 10).}
        \label{fig:cluster_analysis_en}
    \end{subfigure}
    \hfill
    \begin{subfigure}[t]{0.49\linewidth}
        \centering
        \includegraphics[width=\linewidth]{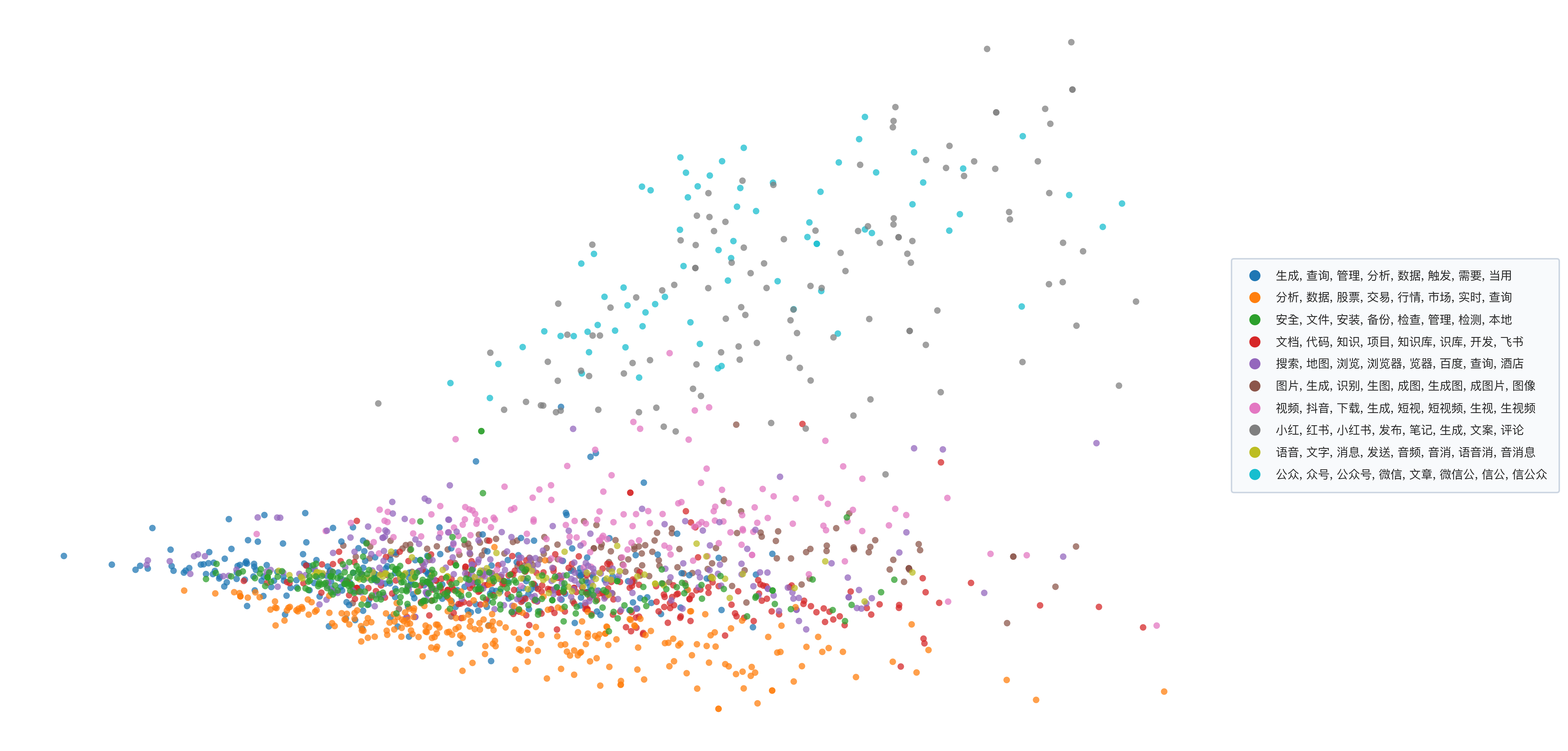}
        \caption{Chinese skills (number of clusters = 10).}
        \label{fig:cluster_analysis_zh}
    \end{subfigure}
    \caption{Cluster analysis.}
    \label{fig:cluster_analysis}
\end{figure}

Furthermore, we conducted a clustering analysis over skill functionality. Given the substantial differences between Chinese and English tokenization, we analyzed Chinese and English skills separately. After filtering out low-quality skills with unclear functionality or missing descriptions, we retained 17,499 English skills and 3,882 Chinese skills for downstream analysis. For each skill, we concatenated its title and summary as a concise function-oriented textual representation, and then applied language-specific tokenization. We further removed non-functional and template-like tokens, such as personal pronouns, prepositions, articles, and other high-frequency words that contribute little to semantic discrimination. Based on these cleaned textual representations, we constructed TF-IDF feature vectors for each language subset, reduced the feature space with Truncated SVD, and then applied K-means clustering to identify major functional groups of skills. To improve interpretability and cross-language comparability, we fixed the number of clusters to ten for each language while additionally examining elbow and silhouette scores as auxiliary references for cluster structure.

As shown in Figure~\ref{fig:cluster_analysis_en} and ~\ref{fig:cluster_analysis_zh}, the comparison between the English and Chinese scatter plots reveals a notable cross-lingual difference in the geometric organization of skill clusters. The English skills are more concentrated in the main semantic region, where most clusters are compressed into a relatively compact wedge-shaped area, despite the presence of one prominent elongated branch. In contrast, the Chinese skills are distributed over a broader area, with several clusters extending upward and separating into more visibly detached regions. This pattern suggests a structural difference in ecosystem orientation. The English skill space appears to be shaped more strongly by technical and infrastructure-oriented capabilities, such as API interaction, code generation, browser automation, data processing, and system-level agent support. Because these capabilities function as reusable modules that can participate in many downstream workflows, their semantic boundaries are often less distinct, leading to stronger overlap among clusters. By contrast, the Chinese skill space appears to be more application-driven, with many skills centered on concrete end-user scenarios such as media production, video generation, social-media operations, and finance-related services. These application-oriented skills usually have clearer task scopes and more explicit functional identities, which in turn yields better cluster separability. Overall, the contrast indicates that English skills are more aligned with capability modularization, whereas Chinese skills are more aligned with scenario-specific packaging.

\begin{figure}[htbp]
\centering
\includegraphics[width=1.0\linewidth]{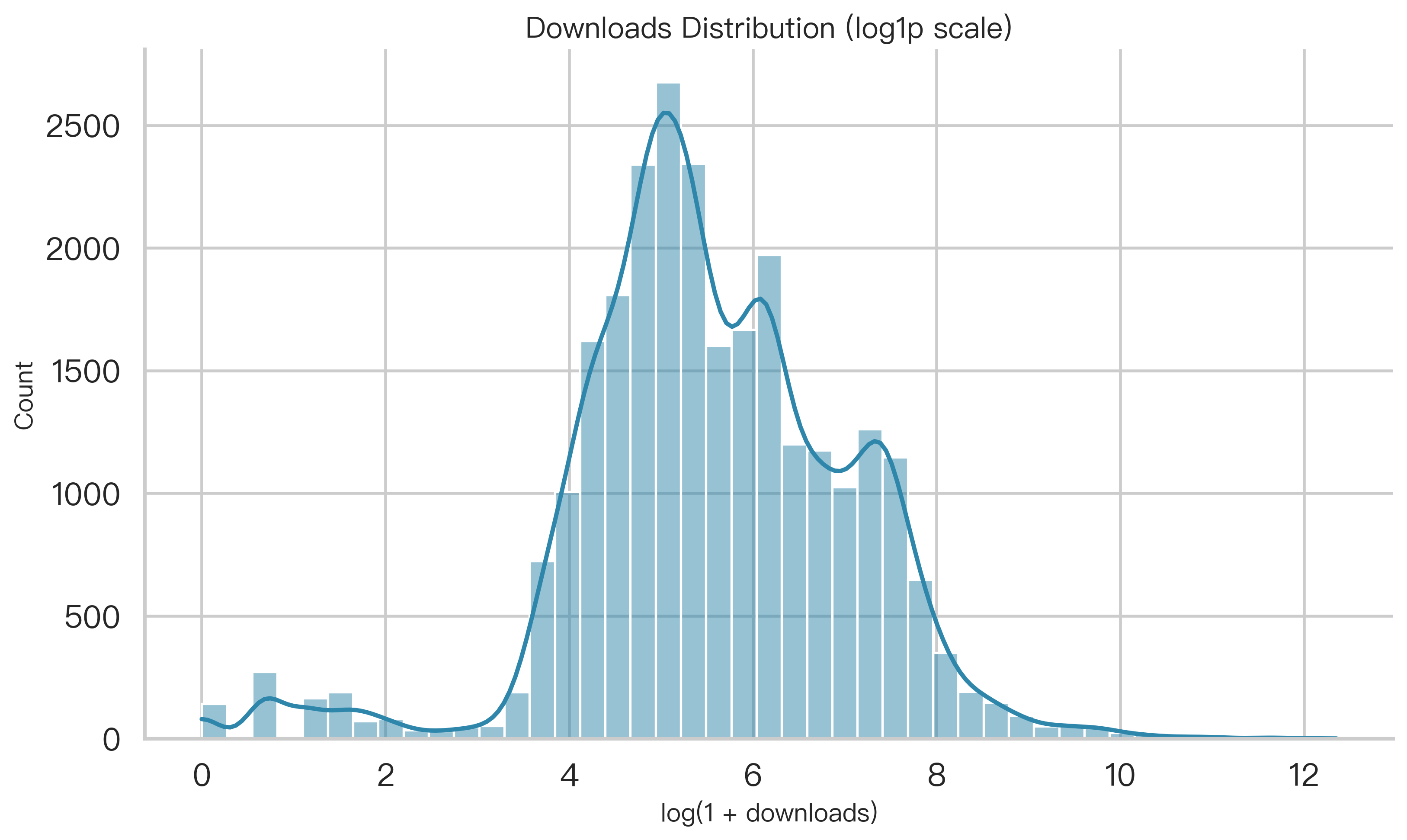}
\caption{Download distribution of skills.}
\label{fig:download_distribution}
\end{figure}

In addition, we analyzed the distribution of skill downloads, as shown in Figure~\ref{fig:download_distribution}. The resulting distribution is approximately normal, with the majority of skills concentrated in the range of 30 to 60 downloads. Only a small fraction of skills, less than 10\%, exceed 1,000 downloads. This suggests that the skill ecosystem exhibits a pronounced long-tail popularity structure: most skills receive only modest attention, while a small fraction of highly downloaded skills account for a disproportionate share of adoption and ecosystem visibility.

\subsection{Risk Analysis}
\begin{figure}[htbp]
    \centering
    \begin{subfigure}[t]{0.55\linewidth}
        \centering
        \includegraphics[width=\linewidth]{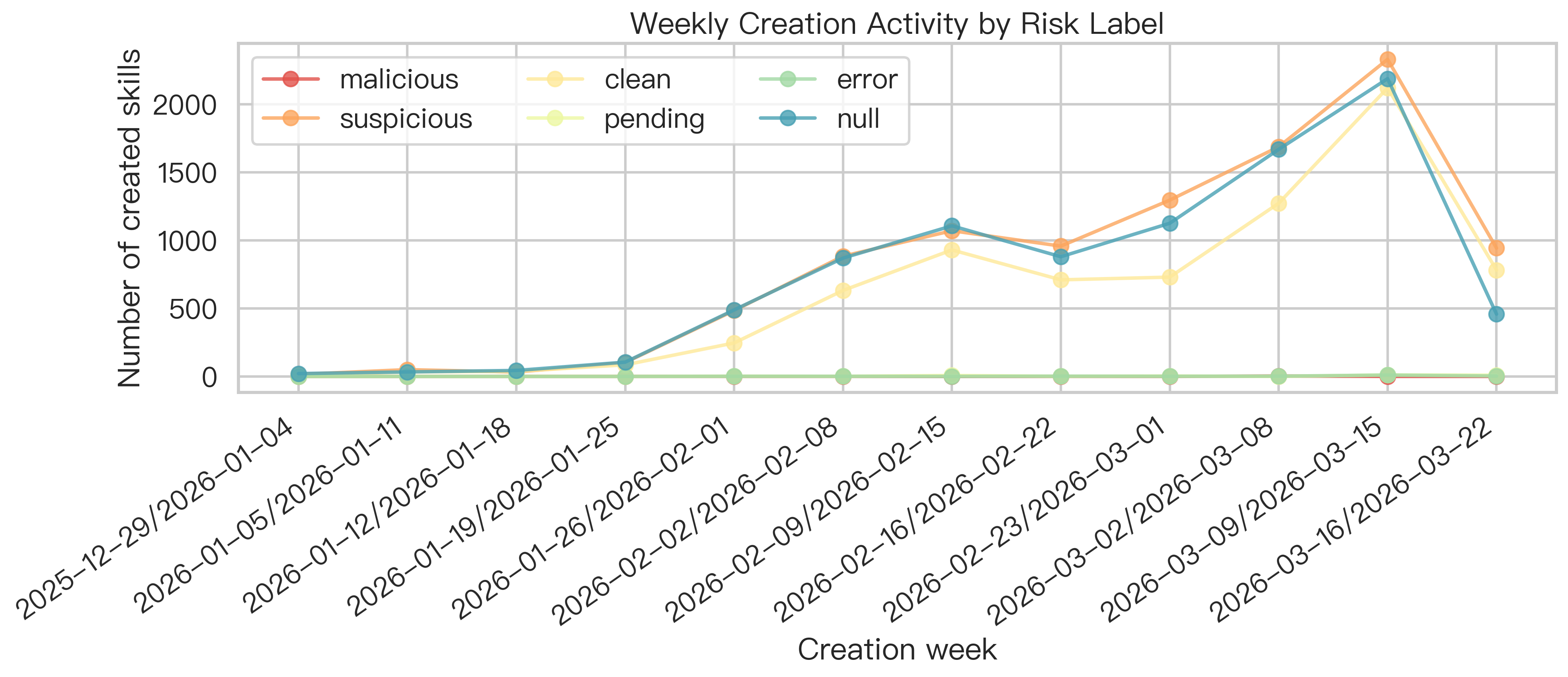}
        \caption{Weekly risk activity.}
        \label{fig:weekly_risk_activity}
    \end{subfigure}
    \hfill
    \begin{subfigure}[t]{0.43\linewidth}
        \centering
        \includegraphics[width=\linewidth]{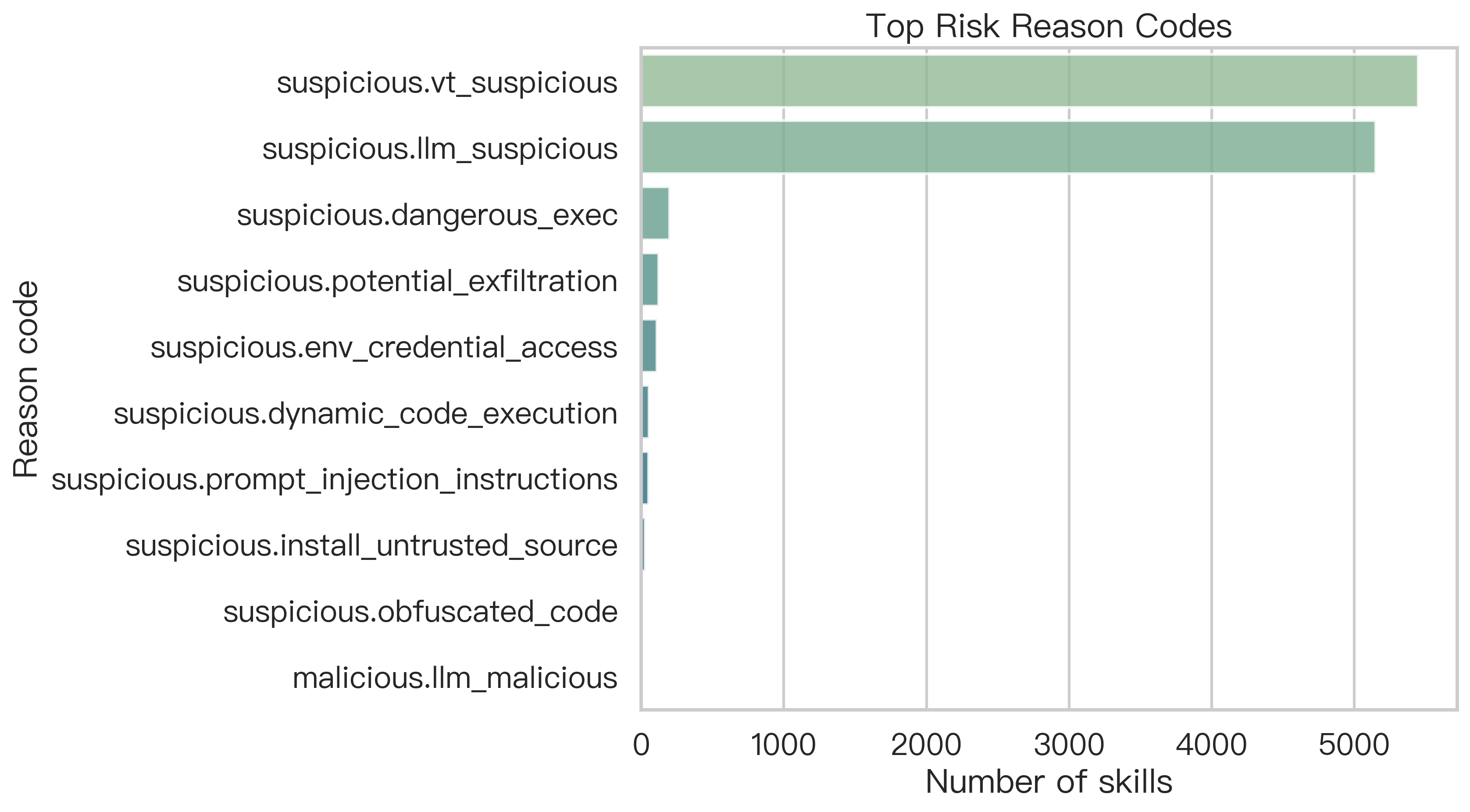}
        \caption{Distribution of different security risk categories.}
        \label{fig:security_risk_distribution}
    \end{subfigure}
    \label{fig:risk_overview}
    \caption{Risk overview.}
\end{figure}

In addition to the functional analysis, we conducted an in-depth assessment of skill risk profiles using publicly available tags on ClawHub related to risk scanning, evaluation, and relevant labels. As shown in Figures~\ref{fig:weekly_risk_activity} and ~\ref{fig:security_risk_distribution}, we first quantified the temporal trends and categorical distributions of risk, performing statistical analyses across two dimensions: time and risk category.

As shown in Figure~\ref{fig:weekly_risk_activity}, from the temporal perspective, despite the surge in the number of skills, the proportion of suspicious skills has not declined and has remained relatively stable. This suggests that security within the current skills community remains a significant unresolved risk. Furthermore, the number of skills that are not labeled as either safe or suspicious has also stayed persistently high, indicating substantial gaps in the community’s security oversight and monitoring mechanisms.

\begin{figure}[htbp]
\centering
\includegraphics[width=0.7\linewidth]{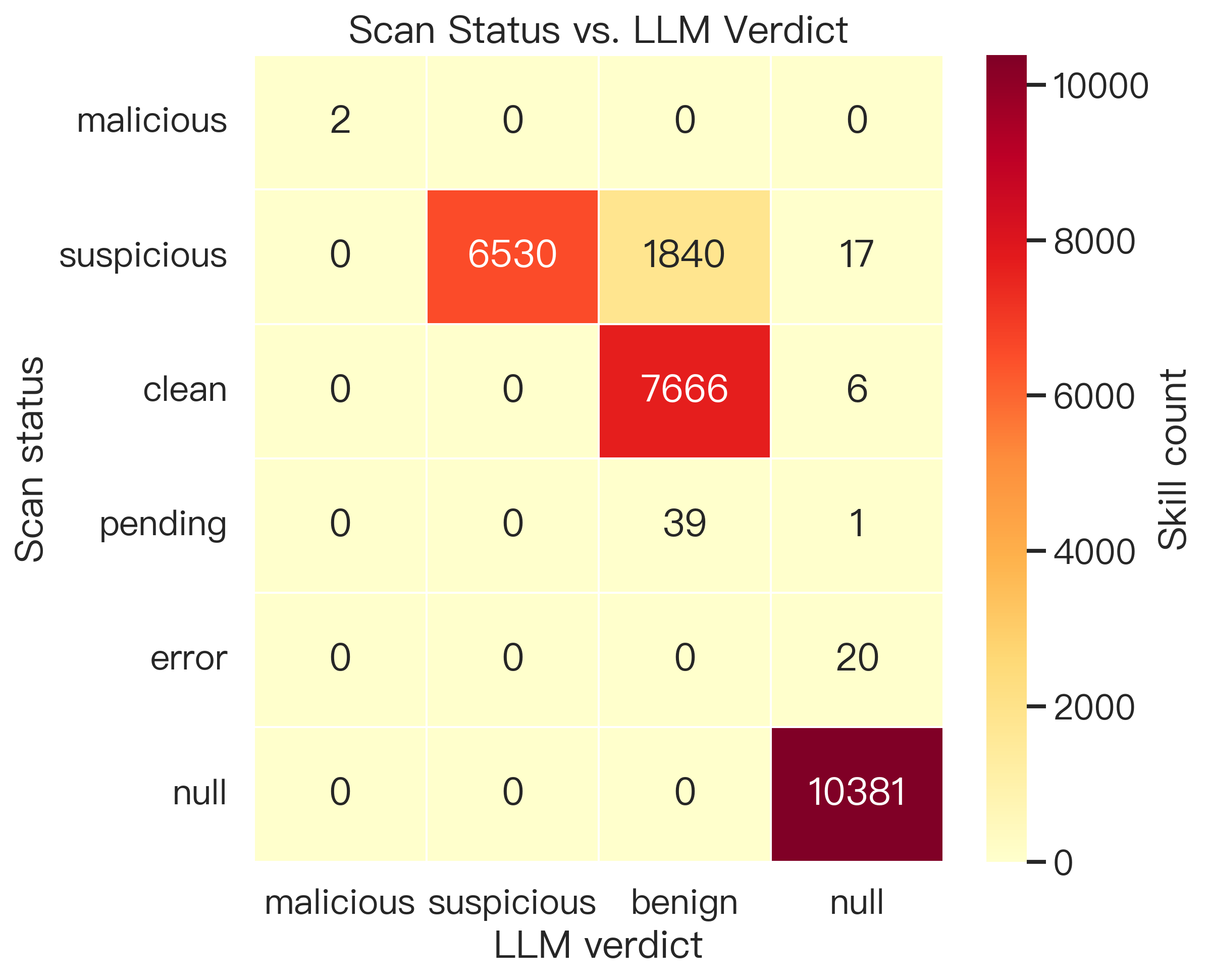}
\caption{Comparison between risk detection by LLMs and security scanning tools.}
\label{fig:security_scan_comparison}
\end{figure}

From the risk-label distribution shown in Figure~\ref{fig:security_risk_distribution}, most risk annotations originate from the VT scanning tool and from LLM-as-a-Judge. While judgments produced by large models have some informative value, as shown in Figure~\ref{fig:security_scan_comparison} the model tends to be conservative: among all vulnerabilities detected by VT, the model classifies approximately 20\% of those samples as benign. In addition, \textbf{fine-grained labels remain severely lacking}. Of the 9,000+ suspicious skills, only about 600 have specific risk labels. The existing risk labels are grouped into seven categories, and for four of those categories the sample counts are below 100, which poses a significant data challenge for training risk detection models.

\begin{figure}[htbp]
    \centering
    \begin{subfigure}[t]{0.49\linewidth}
        \centering
        \includegraphics[width=\linewidth]{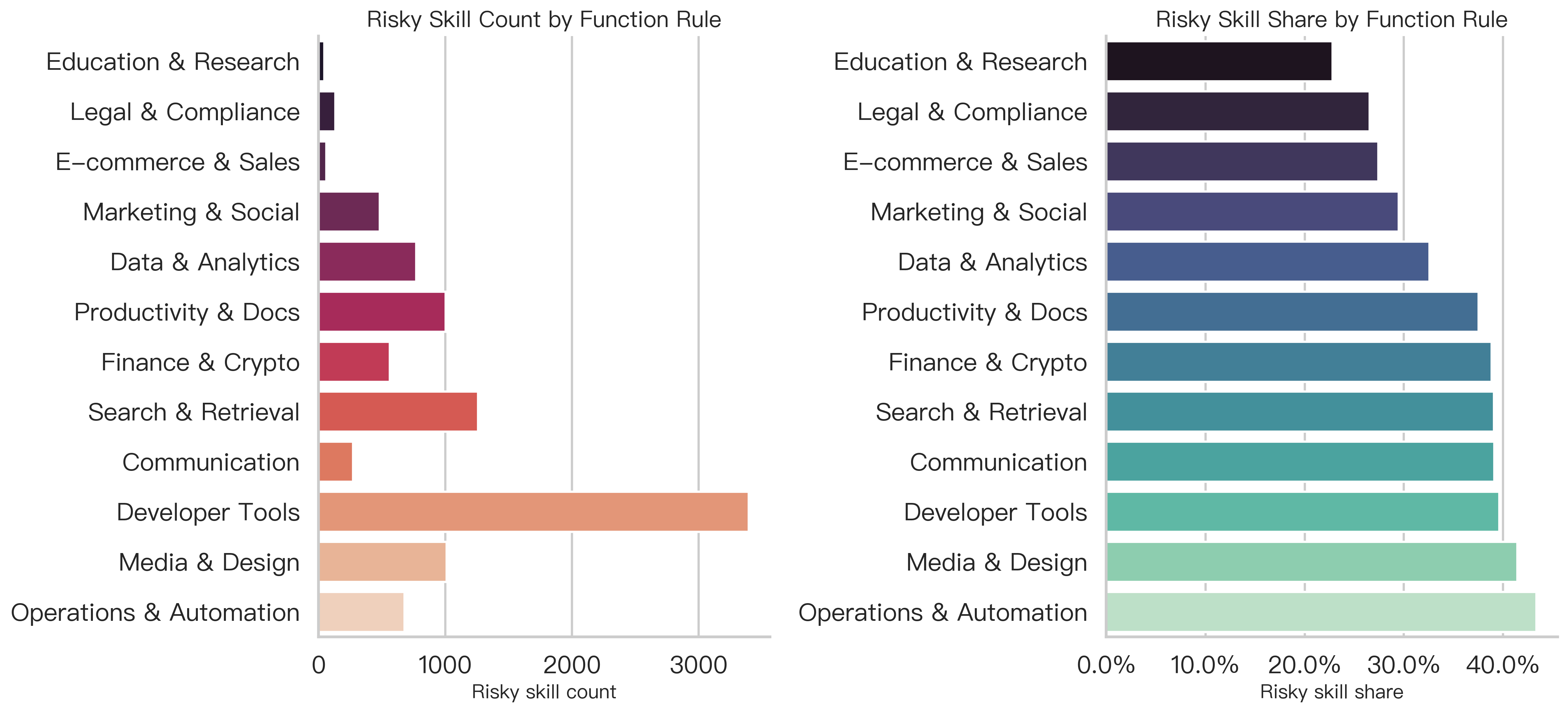}
        \caption{Risk distribution by skill domain.}
        \label{fig:risk_distribution_by_domain}
    \end{subfigure}
    \hfill
    \begin{subfigure}[t]{0.49\linewidth}
        \centering
        \includegraphics[width=\linewidth]{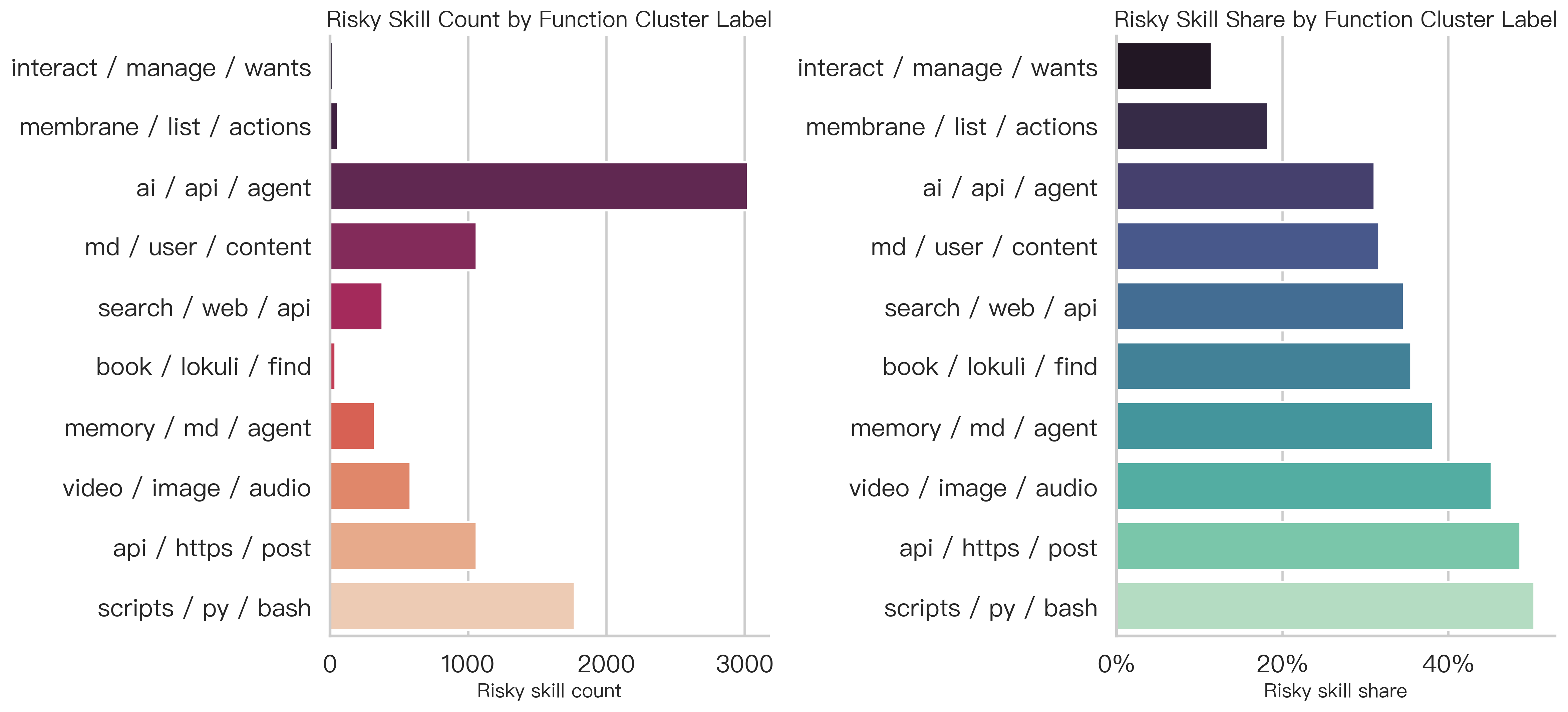}
        \caption{Risk distribution by skill function.}
        \label{fig:risk_distribution_by_function}
    \end{subfigure}
    \caption{Risk distribution by domain and function.}
\end{figure}

Building on this, we further investigated which types of skills are more likely to carry risks. As shown in Figures~\ref{fig:risk_distribution_by_domain} and ~\ref{fig:risk_distribution_by_function}, we visualized the risk distribution of skills by domain and by functionality.

From the visualization we observe that, developer-focused and script-related skills (e.g., “developer tools”, “scripts / py / bash”, and “ai / api / agent”) account for the largest absolute counts of risky skills, while domains like Operations \& Automation, Media \& Design, and Productivity show high risky-share percentages. Functionally, capabilities that execute code, perform outbound network/API calls, or process multimedia are disproportionately associated with risk—these features increase attack surface through command execution, data exfiltration, or unsafe third‑party interactions.

To mitigate these risks, platforms should prioritize scrutiny of high‑risk functionality clusters by enforcing least‑privilege capability gating (sandboxing, scoped tokens, allowlists), expanding targeted labeling and active sampling for underrepresented risk types, and deploying runtime behavioral monitoring to detect malicious behavior that static checks miss. These measures help reduce prevalence and improve detection.

\begin{figure}[htbp]
\centering
\includegraphics[width=1.0\linewidth]{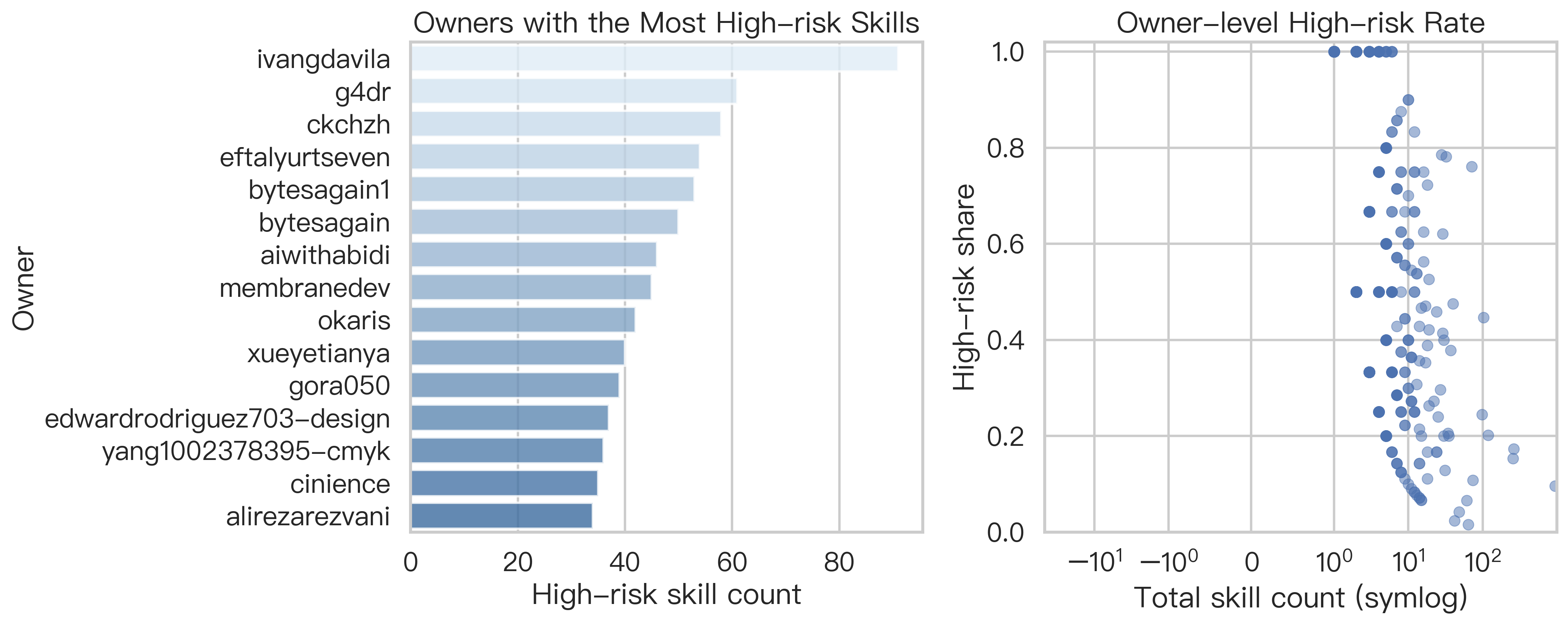}
\caption{List of owners aggregating high-risk skills.}
\label{fig:high_risk_skill_owner}
\end{figure}

To understand the sources of these risks, as shown in Figure~\ref{fig:high_risk_skill_owner}, we analyzed the top 15 contributors of risky skills and found that the number of skills contributed by an author correlates closely with the number of risky skills they produce. Many of these top contributors are companies or organizations that are not acting maliciously toward the platform or users; rather, the risks stem from the complexity and usage scenarios of their skills. This highlights the complexity of risk monitoring: beyond guarding against hackers and bad actors, certain behaviors enabled by skills, especially in contexts like automation and programming, are intrinsically risky and prone to exploitation in production. Even skills that are benign at upload can have vulnerabilities discovered later. Consequently, long‑term tracking and management of risk is essential and requires continuous effort and improvement from the community.

\subsection{Risk Detection}
To evaluate current techniques for detecting skill risk, we conducted a preliminary experiment. We selected twelve mainstream classification methods and used submission-time information (including descriptions, documentation, and code) as input features to predict whether a skill is risky. Specifically, we filtered 11,010 records that had both risk labels and sufficient basic information for classification, and split them randomly into training and test sets at an 8:1 ratio, yielding 8,808 training samples and 2,202 test samples. Suspicious skills were treated as positive samples, with an overall positive-to-negative ratio of 1:1.

\begin{table*}[htbp]
  \centering
  \caption{Risk detection results of different classifiers.}
    \begin{tabular}{lrrrrrrr}
    \toprule
    Method & Feature & ACC   & PRE   & REC   & F1    & AUC   & AVG \\
    \midrule
    Logistic Regression & 40910 & 72.62 & 72.68 & 72.48 & 72.58 & 78.95 & 73.86 \\
    MLP   & 128   & 71    & 68.81 & 76.75 & 72.57 & 77.49 & 73.32 \\
    Random Forest & 256   & 70.53 & 68.08 & 77.29 & 72.39 & 77.01 & 73.06 \\
    Extra Trees & 256   & 68.89 & 66.46 & 76.3  & 71.04 & 76.48 & 71.83 \\
    Hist Gradient Boosting & 256   & 69.98 & 68.68 & 73.48 & 70.99 & 77.63 & 72.15 \\
    SGD   & 40910 & 68.94 & 68.9  & 69.03 & 68.97 & 75.19 & 70.21 \\
    Ridge Classifier & 40910 & 68.94 & 68.97 & 68.85 & 68.91 & 75.32 & 70.2 \\
    Linear SVM & 40910 & 68.58 & 68.65 & 68.39 & 68.52 & 74.94 & 69.82 \\
    Naive Bayes & 40910 & 68.21 & 68.42 & 67.67 & 68.04 & 74.17 & 69.3 \\
    Perceptron & 40910 & 67.08 & 67.71 & 65.31 & 66.48 & 73.94 & 68.1 \\
    Adaboost & 128   & 65.31 & 64.85 & 66.85 & 65.83 & 72.02 & 66.97 \\
    KNN   & 128   & 64.67 & 67.57 & 56.41 & 61.49 & 71.25 & 64.28 \\
    \bottomrule
    \end{tabular}%
  \label{tab:risk_detect}%
\end{table*}%

As shown in Table~\ref{tab:risk_detect}, the results indicate that, despite using relatively simple methods, the classifiers are reasonably effective. Logistic Regression, MLP, and Random Forest all achieve classification accuracies above 70\%, with the best Logistic Regression model reaching 72.62\%. This is less than 8 percentage points below the large-model–based classifier (80\%), demonstrating the viability of feature-based classification. By engineering more detailed features (e.g., whether root privileges are required or whether sensitive data is accessed), classifier-based approaches hold promise for further improving skill-risk detection.

\begin{table}[htbp]
  \centering
  \caption{Ablation of risk detection inputs on Logistic Regression.}
    \begin{tabular}{lrrrrrr}
    \toprule
    Feature & \multicolumn{1}{l}{ACC} & \multicolumn{1}{l}{PRE} & \multicolumn{1}{l}{REC} & \multicolumn{1}{l}{F1} & \multicolumn{1}{l}{AUC} & \multicolumn{1}{l}{AVG} \\
    \midrule
    drop changelog & 70.21 & 69.71 & 71.48 & 70.58 & 78.12 & 72.02 \\
    drop file paths & 70.57 & 70.61 & 70.48 & 70.55 & 78.33 & 72.11 \\
    drop name & 71.48 & 71.33 & 71.84 & 71.58 & 79.12 & 73.07 \\
    drop structure & 71.66 & 71.43 & 72.21 & 71.82 & 78.29 & 73.08 \\
    drop summary & 72.84 & 72.68 & 73.21 & 72.94 & 78.91 & 74.12 \\
    drop tags & 72.25 & 72.31 & 72.12 & 72.21 & 78.87 & 73.55 \\
    drop primary doc & 70.07 & 70.13 & 69.94 & 70.03 & 76.57 & 71.35 \\
    full  & 72.62 & 72.68 & 72.48 & 72.58 & 78.95 & 73.86 \\
    \bottomrule
    \end{tabular}%
  \label{tab:ablation}%
\end{table}%

Building on this, we performed an ablation study on the best-performing model, Logistic Regression, by removing each type of input information in turn and comparing risk-detection performance. As shown in Table~\ref{tab:ablation}, the primary documentation (primary doc) has the largest and most important effect on performance. Besides, removing any information except the summary leads to a notable drop in detection performance; however, removing the summary actually increases accuracy by 0.22\% percentage points. This suggests that summaries contain substantial noise and that some risky skills use seemingly harmless summaries to mislead classifiers, causing misclassification.

\section{Related Work}

The recent rise of LLM-based agents has shifted research attention from static language generation to autonomous systems that can plan, reason, call tools, and interact with external environments. Early works such as ReAct \citep{yao2022react}, Toolformer \citep{schick2023toolformer}, and ART \citep{paranjape2023art} established the basic paradigm of combining language reasoning with external actions and tool invocation. Subsequent studies further expanded the agent setting to more complex and persistent environments, including open-ended embodied skill acquisition in Voyager \citep{wang2023voyager} and long-context memory management in MemGPT \citep{packer2023memgpt}. These works collectively demonstrate that agent capability increasingly depends on modular, composable, and reusable external resources rather than on parametric knowledge alone.

A parallel line of research has focused on agent architectures and coordination mechanisms. Multi-agent collaboration frameworks such as CAMEL \citep{li2023camel}, AutoGen \citep{wu2024autogen}, and MetaGPT \citep{hong2023metagpt} show that complex tasks can be decomposed into interacting roles, conversations, and structured workflows. More broadly, recent surveys have systematized the methodological landscape of LLM agents, covering planning, memory, tool use, communication, and evolutionary mechanisms \citep{qu2025tool,guo2024large,luo2025llmagent}. This literature has clarified how agent systems are built and scaled, but it is still primarily centered on the design of agents themselves rather than on the public ecosystems through which reusable capabilities are packaged and disseminated.

Another growing body of work studies the evaluation of LLM agents in realistic and interactive environments. Benchmarks such as AgentBench \citep{liu2023agentbench} and WebArena \citep{zhou2024webarena} highlight the gap between promising agent abstractions and robust real-world task execution. Recent surveys on agent evaluation further emphasize the need to assess not only capability, but also robustness, safety, efficiency, and long-horizon reliability \citep{yehudai2025evaluation,luo2025llmagent}. Our work is complementary to this literature. Rather than evaluating how well an agent performs after deployment, we focus on the ecosystem layer that precedes deployment: publicly shared skills. In this sense, our study extends agent research from model- and system-level capability analysis toward registry-level measurement and security assessment, asking how reusable agent capabilities are distributed, adopted, and exposed to risk in open skill marketplaces.

\section{Conclusion}
In this paper, we presented a data-driven study of the emerging skill ecosystem through the lens of ClawHub. By constructing a normalized corpus of 26,502 skills, we examined how skills are distributed across languages, functionality, popularity, and risk. Our analysis shows that public skill registries are not merely repositories of reusable agent components; they are evolving into socially distributed software ecosystems with distinct functional structures and nontrivial security implications. In particular, we observed clear cross-lingual differences in the functional composition of skills: English skills are more heavily oriented toward technical and infrastructural capabilities, whereas Chinese skills are more strongly organized around downstream application scenarios such as media, social platforms, and finance-related workflows.
Beyond ecosystem profiling, we showed that risk is already a prominent property of public skill registries. A considerable portion of skills are flagged as suspicious or malicious by available platform signals, yet many records still suffer from incomplete observability, suggesting that the current safety pipeline is informative but far from comprehensive. To move toward practical mitigation, we introduced the task of submission-time skill risk prediction and demonstrated that meaningful predictive performance can be achieved using only the information visible at release time. Our benchmark results further indicate that documentation quality and file-level submission signals are more valuable than short summaries, highlighting the importance of rich submission metadata for early-stage governance.
Overall, our study suggests that skill ecosystems should be treated not only as capability-sharing infrastructures, but also as socio-technical systems that require measurement, auditing, and risk-aware governance. Future work can extend this direction by incorporating longitudinal updates, dynamic execution traces, downstream installation behavior, and cross-registry comparisons, as well as by developing finer-grained models for reason-code prediction, trust calibration, and intervention prioritization.

\bibliographystyle{named}
\bibliography{reference}

@article{bahak2023evaluating,
  title={Evaluating chatgpt as a question answering system: A comprehensive analysis and comparison with existing models},
  author={Bahak, Hossein and Taheri, Farzaneh and Zojaji, Zahra and Kazemi, Arefeh},
  journal={arXiv preprint arXiv:2312.07592},
  year={2023}
}

@inproceedings{wu2023towards,
  title={Towards improving the reliability and transparency of ChatGPT for educational question answering},
  author={Wu, Yongchao and Henriksson, Aron and Duneld, Martin and Nouri, Jalal},
  booktitle={European conference on technology enhanced learning},
  pages={475--488},
  year={2023},
  organization={Springer}
}

@inproceedings{chan2023case,
  title={A case study on chatgpt question generation},
  author={Chan, Winston and An, Aijun and Davoudi, Heidar},
  booktitle={2023 IEEE International Conference on Big Data (BigData)},
  pages={1647--1656},
  year={2023},
  organization={IEEE}
}

@article{shen2023chatgpt,
  title={In chatgpt we trust? measuring and characterizing the reliability of chatgpt},
  author={Shen, Xinyue and Chen, Zeyuan and Backes, Michael and Zhang, Yang},
  journal={arXiv preprint arXiv:2304.08979},
  year={2023}
}

@article{xi2025rise,
  title={The rise and potential of large language model based agents: A survey},
  author={Xi, Zhiheng and Chen, Wenxiang and Guo, Xin and He, Wei and Ding, Yiwen and Hong, Boyang and Zhang, Ming and Wang, Junzhe and Jin, Senjie and Zhou, Enyu and others},
  journal={Science China Information Sciences},
  volume={68},
  number={2},
  pages={121101},
  year={2025},
  publisher={Springer}
}

@article{yang2025comprehensive,
  title={A Comprehensive Survey on Large Language Model based Agents for Education},
  author={Yang, Juan and Wang, Minjuan and Du, Xu and Na, Rina},
  journal={IEEE Transactions on Learning Technologies},
  year={2025},
  publisher={IEEE}
}

@article{guo2024large,
  title={Large language model based multi-agents: A survey of progress and challenges},
  author={Guo, Taicheng and Chen, Xiuying and Wang, Yaqi and Chang, Ruidi and Pei, Shichao and Chawla, Nitesh V and Wiest, Olaf and Zhang, Xiangliang},
  journal={arXiv preprint arXiv:2402.01680},
  year={2024}
}

@article{qin2024tool,
  title={Tool learning with foundation models},
  author={Qin, Yujia and Hu, Shengding and Lin, Yankai and Chen, Weize and Ding, Ning and Cui, Ganqu and Zeng, Zheni and Zhou, Xuanhe and Huang, Yufei and Xiao, Chaojun and others},
  journal={ACM Computing Surveys},
  volume={57},
  number={4},
  pages={1--40},
  year={2024},
  publisher={ACM New York, NY}
}

@article{gou2023tora,
  title={Tora: A tool-integrated reasoning agent for mathematical problem solving},
  author={Gou, Zhibin and Shao, Zhihong and Gong, Yeyun and Shen, Yelong and Yang, Yujiu and Huang, Minlie and Duan, Nan and Chen, Weizhu},
  journal={arXiv preprint arXiv:2309.17452},
  year={2023}
}

@article{singh2025agentic,
  title={Agentic reasoning and tool integration for llms via reinforcement learning},
  author={Singh, Joykirat and Magazine, Raghav and Pandya, Yash and Nambi, Akshay},
  journal={arXiv preprint arXiv:2505.01441},
  year={2025}
}

@article{ling2026agent,
  title={Agent Skills: A Data-Driven Analysis of Claude Skills for Extending Large Language Model Functionality},
  author={Ling, George and Zhong, Shanshan and Huang, Richard},
  journal={arXiv preprint arXiv:2602.08004},
  year={2026}
}

@article{li2026skillsbench,
  title={SkillsBench: Benchmarking how well agent skills work across diverse tasks},
  author={Li, Xiangyi and Chen, Wenbo and Liu, Yimin and Zheng, Shenghan and Chen, Xiaokun and He, Yifeng and Li, Yubo and You, Bingran and Shen, Haotian and Sun, Jiankai and others},
  journal={arXiv preprint arXiv:2602.12670},
  year={2026}
}

@article{chen2026skills,
  title={Skills Are the New Apps--NowIt’s Time for Skill OS},
  author={Chen, Le and Wang, Zichang and Zheng, Wenxin and Feng, Erhu and Du, Dong and Xia, Yubin and Chen, Haibo},
  year={2026},
  publisher={Preprints}
}

@article{wang2025prompt,
  title={Prompt engineering for healthcare: Methodologies and applications},
  author={Wang, Jiaqi and Shi, Enze and Yu, Sigang and Wu, Zihao and Hu, Huawen and Ma, Chong and Dai, Haixing and Yang, Qiushi and Kang, Yanqing and Wu, Jinru and others},
  journal={Meta-Radiology},
  pages={100190},
  year={2025},
  publisher={Elsevier}
}

@article{giray2023prompt,
  title={Prompt engineering with ChatGPT: a guide for academic writers},
  author={Giray, Louie},
  journal={Annals of biomedical engineering},
  volume={51},
  number={12},
  pages={2629--2633},
  year={2023},
  publisher={Springer}
}

@article{panda2024revolutionizing,
  title={Revolutionizing language processing in libraries with SheetGPT: an integration of Google Sheet and ChatGPT plugin},
  author={Panda, Subhajit and Kaur, Navkiran},
  journal={Library Hi Tech News},
  volume={41},
  number={5},
  pages={8--10},
  year={2024},
  publisher={Emerald Publishing Limited}
}

@article{liu2024toolace,
  title={Toolace: Winning the points of llm function calling},
  author={Liu, Weiwen and Huang, Xu and Zeng, Xingshan and Hao, Xinlong and Yu, Shuai and Li, Dexun and Wang, Shuai and Gan, Weinan and Liu, Zhengying and Yu, Yuanqing and others},
  journal={arXiv preprint arXiv:2409.00920},
  year={2024}
}

@article{wang2025function,
  title={Function Calling in Large Language Models: Industrial Practices, Challenges, and Future Directions},
  author={Wang, Maolin and Zhang, Yingyi and Yu, Bowen and Hao, Bingguang and Peng, Cunyin and Chen, Yicheng and Zhou, Wei and Gu, Jinjie and Zhuang, Chenyi and Guo, Ruocheng and others},
  journal={ACM Computing Surveys},
  year={2025},
  publisher={ACM New York, NY}
}

@inproceedings{lu2025toolsandbox,
  title={Toolsandbox: A stateful, conversational, interactive evaluation benchmark for llm tool use capabilities},
  author={Lu, Jiarui and Holleis, Thomas and Zhang, Yizhe and Aumayer, Bernhard and Nan, Feng and Bai, Haoping and Ma, Shuang and Ma, Shen and Li, Mengyu and Yin, Guoli and others},
  booktitle={Findings of the Association for Computational Linguistics: NAACL 2025},
  pages={1160--1183},
  year={2025}
}

@article{shen2024llm,
  title={Llm with tools: A survey},
  author={Shen, Zhuocheng},
  journal={arXiv preprint arXiv:2409.18807},
  year={2024}
}

@article{wang2025mcp,
  title={Mcp-bench: Benchmarking tool-using llm agents with complex real-world tasks via mcp servers},
  author={Wang, Zhenting and Chang, Qi and Patel, Hemani and Biju, Shashank and Wu, Cheng-En and Liu, Quan and Ding, Aolin and Rezazadeh, Alireza and Shah, Ankit and Bao, Yujia and others},
  journal={arXiv preprint arXiv:2508.20453},
  year={2025}
}

@article{ahmadi2025mcp,
  title={Mcp bridge: A lightweight, llm-agnostic restful proxy for model context protocol servers},
  author={Ahmadi, Arash and Sharif, Sarah and Banad, Yaser M},
  journal={arXiv preprint arXiv:2504.08999},
  year={2025}
}

@article{sarkar2025survey,
  title={Survey of llm agent communication with mcp: A software design pattern centric review},
  author={Sarkar, Anjana and Sarkar, Soumyendu},
  journal={arXiv preprint arXiv:2506.05364},
  year={2025}
}

@article{liang2026skillnet,
  title={SkillNet: Create, Evaluate, and Connect AI Skills},
  author={Liang, Yuan and Zhong, Ruobin and Xu, Haoming and Jiang, Chen and Zhong, Yi and Fang, Runnan and Gu, Jia-Chen and Deng, Shumin and Yao, Yunzhi and Wang, Mengru and others},
  journal={arXiv preprint arXiv:2603.04448},
  year={2026}
}

@article{he2026openclaw,
  title={OpenClaw as Language Infrastructure: A Case-Centered Survey of a Public Agent Ecosystem in the Wild},
  author={He, Chaoyue and Zhou, Xin and Wang, Di and Xu, Hong and Liu, Wei and Miao, Chunyan},
  year={2026},
  publisher={Preprints}
}

@online{anthropic2025lifesciences,
  author       = {{Anthropic}},
  title        = {Claude for Life Sciences},
  year         = {2025},
  month        = oct,
  day          = {20},
  url          = {https://www.anthropic.com/news/claude-for-life-sciences},
  note         = {Accessed: 2026-03-19}
}

@online{openclaw2026skills,
  author       = {{OpenClaw}},
  title        = {Skills},
  year         = {2026},
  url          = {https://docs.openclaw.ai/skills},
  note         = {OpenClaw documentation. Accessed: 2026-03-19}
}

@online{openclaw2026clawhub,
  author       = {{OpenClaw}},
  title        = {ClawHub},
  year         = {2026},
  url          = {https://docs.openclaw.ai/tools/clawhub},
  note         = {OpenClaw documentation. Accessed: 2026-03-19}
}

@online{theverge2026openclaw,
  author       = {{The Verge}},
  title        = {OpenClaw's AI 'skill' extensions are a security nightmare},
  year         = {2026},
  month        = feb,
  url          = {https://www.theverge.com/news/874011/openclaw-ai-skill-clawhub-extensions-security-nightmare},
  note         = {Accessed: 2026-03-19}
}

@online{tomshardware2026clawhub,
  author       = {{Tom's Hardware}},
  title        = {Malicious OpenClaw 'skill' targets crypto users on ClawHub},
  year         = {2026},
  month        = feb,
  url          = {https://www.tomshardware.com/tech-industry/cyber-security/malicious-moltbot-skill-targets-crypto-users-on-clawhub},
  note         = {Accessed: 2026-03-19}
}

@inproceedings{yao2022react,
  title={React: Synergizing reasoning and acting in language models},
  author={Yao, Shunyu and Zhao, Jeffrey and Yu, Dian and Du, Nan and Shafran, Izhak and Narasimhan, Karthik R and Cao, Yuan},
  booktitle={The eleventh international conference on learning representations},
  year={2022}
}

@article{schick2023toolformer,
  title={Toolformer: Language models can teach themselves to use tools},
  author={Schick, Timo and Dwivedi-Yu, Jane and Dess{\`\i}, Roberto and Raileanu, Roberta and Lomeli, Maria and Hambro, Eric and Zettlemoyer, Luke and Cancedda, Nicola and Scialom, Thomas},
  journal={Advances in neural information processing systems},
  volume={36},
  pages={68539--68551},
  year={2023}
}

@article{paranjape2023art,
  title={Art: Automatic multi-step reasoning and tool-use for large language models},
  author={Paranjape, Bhargavi and Lundberg, Scott and Singh, Sameer and Hajishirzi, Hannaneh and Zettlemoyer, Luke and Ribeiro, Marco Tulio},
  journal={arXiv preprint arXiv:2303.09014},
  year={2023}
}

@article{wang2023voyager,
  title={Voyager: An open-ended embodied agent with large language models},
  author={Wang, Guanzhi and Xie, Yuqi and Jiang, Yunfan and Mandlekar, Ajay and Xiao, Chaowei and Zhu, Yuke and Fan, Linxi and Anandkumar, Anima},
  journal={arXiv preprint arXiv:2305.16291},
  year={2023}
}

@article{packer2023memgpt,
  title={MemGPT: towards LLMs as operating systems.},
  author={Packer, Charles and Fang, Vivian and Patil, Shishir\_G and Lin, Kevin and Wooders, Sarah and Gonzalez, Joseph\_E},
  year={2023},
  publisher={ArXiv}
}

@article{li2023camel,
  title={Camel: Communicative agents for" mind" exploration of large language model society},
  author={Li, Guohao and Hammoud, Hasan and Itani, Hani and Khizbullin, Dmitrii and Ghanem, Bernard},
  journal={Advances in neural information processing systems},
  volume={36},
  pages={51991--52008},
  year={2023}
}

@inproceedings{wu2024autogen,
  title={Autogen: Enabling next-gen LLM applications via multi-agent conversations},
  author={Wu, Qingyun and Bansal, Gagan and Zhang, Jieyu and Wu, Yiran and Li, Beibin and Zhu, Erkang and Jiang, Li and Zhang, Xiaoyun and Zhang, Shaokun and Liu, Jiale and others},
  booktitle={First conference on language modeling},
  year={2024}
}

@inproceedings{hong2023metagpt,
  title={MetaGPT: Meta programming for a multi-agent collaborative framework},
  author={Hong, Sirui and Zhuge, Mingchen and Chen, Jonathan and Zheng, Xiawu and Cheng, Yuheng and Wang, Jinlin and Zhang, Ceyao and Wang, Zili and Yau, Steven Ka Shing and Lin, Zijuan and others},
  booktitle={The twelfth international conference on learning representations},
  year={2023}
}

@article{qu2025tool,
  title={Tool learning with large language models: A survey},
  author={Qu, Changle and Dai, Sunhao and Wei, Xiaochi and Cai, Hengyi and Wang, Shuaiqiang and Yin, Dawei and Xu, Jun and Wen, Ji-Rong},
  journal={Frontiers of Computer Science},
  volume={19},
  number={8},
  pages={198343},
  year={2025},
  publisher={Springer}
}

@article{luo2025llmagent,
  title         = {Large Language Model Agent: A Survey on Methodology, Applications and Challenges},
  author        = {Luo, Junyu and Zhang, Weizhi and Yuan, Ye and Zhao, Yusheng and Yang, Junwei and Gu, Yiyang and Wu, Bohan and Chen, Binqi and Qiao, Ziyue and Long, Qingqing and Tu, Rongcheng and Luo, Xiao and Ju, Wei and Xiao, Zhiping and Wang, Yifan and Xiao, Meng and Liu, Chenwu and Yuan, Jingyang and Zhang, Shichang and Jin, Yiqiao and Zhang, Fan and Wu, Xian and Zhao, Hanqing and Tao, Dacheng and Yu, Philip S. and Zhang, Ming},
  journal       = {arXiv preprint arXiv:2503.21460},
  year          = {2025},
  doi           = {10.48550/arXiv.2503.21460},
  archivePrefix = {arXiv},
  eprint        = {2503.21460},
  primaryClass  = {cs.CL},
  url           = {https://arxiv.org/abs/2503.21460}
}

@article{liu2023agentbench,
  title         = {AgentBench: Evaluating LLMs as Agents},
  author        = {Liu, Xiao and Yu, Hao and Zhang, Hanchen and Xu, Yifan and Lei, Xuanyu and Lai, Hanyu and Gu, Yu and Ding, Hangliang and Men, Kaiwen and Yang, Kejuan and Zhang, Shudan and Deng, Xiang and Zeng, Aohan and Chen, Huihui and Zhang, Chenyu and Zhang, Tianyang and Su, Yu and Sun, Maosong and Tang, Jie},
  journal       = {arXiv preprint arXiv:2308.03688},
  year          = {2023},
  doi           = {10.48550/arXiv.2308.03688},
  archivePrefix = {arXiv},
  eprint        = {2308.03688},
  primaryClass  = {cs.AI},
  url           = {https://arxiv.org/abs/2308.03688}
}

@article{zhou2024webarena,
  title         = {WebArena: A Realistic Web Environment for Building Autonomous Agents},
  author        = {Zhou, Shuyan and Xu, Frank F. and Zhu, Hao and Zhou, Xuhui and Lo, Robert and Sridhar, Abishek and Cheng, Xianyi and Ou, Tianyue and Bisk, Yonatan and Fried, Daniel and Alon, Uri and Neubig, Graham},
  journal       = {arXiv preprint arXiv:2307.13854},
  year          = {2024},
  doi           = {10.48550/arXiv.2307.13854},
  archivePrefix = {arXiv},
  eprint        = {2307.13854},
  primaryClass  = {cs.AI},
  url           = {https://arxiv.org/abs/2307.13854}
}

@article{yehudai2025evaluation,
  title         = {Survey on Evaluation of LLM-based Agents},
  author        = {Yehudai, Asaf and Eden, Lilach and Li, Alan and Uziel, Guy and Zhao, Yilun and Bar-Haim, Roy and Cohan, Arman and Shmueli-Scheuer, Michal},
  journal       = {arXiv preprint arXiv:2503.16416},
  year          = {2025},
  doi           = {10.48550/arXiv.2503.16416},
  archivePrefix = {arXiv},
  eprint        = {2503.16416},
  primaryClass  = {cs.AI},
  url           = {https://arxiv.org/abs/2503.16416}
}

\end{document}